# Should Social Robots in Retail Manipulate Customers?

**Oliver Bendel and Liliana Margarida Dos Santos Alves**

School of Business FHNW, Bahnhofstrasse 6, CH-5210 Windisch
oliver.bendel@fhnw.ch; alves.liliana1989@gmail.com

## Abstract

Against the backdrop of structural changes in the retail trade, social robots have found their way into retail stores and shopping malls in order to attract, welcome, and greet customers; to inform them, advise them, and persuade them to make a purchase. Salespeople often have a broad knowledge of their product and rely on offering competent and honest advice, whether it be on shoes, clothing, or kitchen appliances. However, some frequently use sales tricks to secure purchases. The question arises of how consulting and sales robots should "behave". Should they behave like human advisors and salespeople, i.e., occasionally manipulate customers? Or should they be more honest and reliable than us? This article tries to answer these questions. After explaining the basics, it evaluates a study in this context and gives recommendations for companies that want to use consulting and sales robots. Ultimately, fair, honest, and trustworthy robots in retail are a win-win situation for all concerned.

## Introduction

A structural change is reshaping the retail trade. Stationary trade is losing ground to e-commerce. There are attempts to regain competitive parity and reverse this development with the use of online retailing through onsite technologies (including sales and consulting or advisor robots). Robots are seen as having great, but as yet untapped, potential. For example, it would be possible to operate 24/7 and make cost savings by substituting personnel. A higher productivity and profit maximization of the stationary trade is conceivable, furthermore better advisory services, since more extensive and more current information can be called up by robots than people. Years ago, social robots such as Pepper, NAO, and Cruzr made their debut to welcome, inform, advise customers, and ultimately persuade them to make a purchase.

Social and, particularly, humanoid robots often function as service robots, but go far beyond classical models (think of cleaning robots for the floor or for windows), in their design, their natural language capabilities, and other functions of artificial intelligence (AI) like face and voice recognition. Switzerland has been a pioneer in this field, with several projects in Zurich and the surrounding area. Other locations involved in this include California, Japan, and Germany where people rely on relevant social robots (Bendel 2021b).

Consultants and salespeople often have broad knowledge of their product and rely on competent and, all in all, honest advice, whether it be on shoes, clothing, toys, or tools. However, some of them frequently use sales tricks and attempt to outsmart customers to secure purchases. They persuade the customer, for instance, that a shirt or blouse looks good on him or her, or that he or she urgently needs a certain kitchen appliance, which may not be the truth in reality. To boost sales, they use certain tricks and strategies they have learned in specialized sales training courses. Over the course of their professional life, additional skills are added.

Several questions arise from this: How should consulting and sales robots "behave"? Should they behave like human advisors and salespeople by occasionally manipulating customers? Or should they instead be more honest and reliable than humans? What do customers really want? Could it be that social robots are conceded more than people when they utter half-truths or deceptions? After all, they have no conscience – which in turn raises the question of the conscience of the (actual responsible) people behind the sales robot.

While there are many studies on how social robots should behave in principle (Bendel 2021a), and some articles on social robots in retail (Aaltonen et al. 2017), how social robots should behave specifically in retail has been rarely examined. This paper explores the question of whether social robots in retail should manipulate customers or be fair, honest, and trustworthy. First, it outlines how social robots are used in retail stores today. Then, their AI-related capabilities are presented. The next section summarizes a study conducted by Liliana Alves as part of her master's thesis. She surveyed over 300 people on whether consulting and sales robots should manipulate customers. Finally, recommendations are given for companies on how to use social robots in retail. The articles rounds off with a summary and outlook.

# Social Robots in Retail

The following section discusses social robots in retail. First, the terms "social robot" and "service robot" are clarified. Then, examples of robots in retail are given. Finally, their functions are discussed, especially AI-based capabilities.

## Social Robots and Service Robots

Social robots are sensorimotor machines created to interact with humans or animals, particularly more sophisticated species (Bendel 2021a). They can be determined through five key aspects. These are: interaction with living beings, communication with living beings, proximity to living beings, representation of (aspects of or features of) living beings (e.g., they have an animaloid or a humanoid design or natural language abilities), and fundamentally, utility for living beings. A broad definition covers software robots as well as hardware robots, and so could include certain chatbots and voice assistants, relativizing the sensorimotor aspect. Social relationships are often one-to-many relationships, not just one-to-one. In retail, both occur – but the social robot can usually only address one person for technical reasons. This can make the others present, such as the customer's partner or friend, feel uncomfortable.

Some social robots are service robots, that is, they handle certain services and provide certain assistance, and conversely, some service robots are also social robots, insofar as they have communication and interaction functions. Typical examples in this intersection are care and therapy robots, but also advisor and sales robots in the retail trade. In this paper, they are seen as social robots that offer certain service functions such as informing and advising a customer or processing their purchase (Meyer et al. 2020a).

Social robotics crosses over with machine ethics. Moral and immoral conversational agents have emerged from this discipline in recent years – including a chatbot that can systematically lie (Bendel et al. 2017).

## Examples for Robots in Retail

Robots of all kinds are appearing in the retail industry (Kelly 2020; Meyer et al. 2020a/b). Some are transport robots like Relay from Savioke, which moves goods through the aisles of a hardware store, others are security robots like K5 from Knightscope, which patrols the grounds of companies and alliances (see Fig. 1). Tory was "employed" as an inventory assistant at Adler Modemärkte AG in Germany (Bendel 2021b). Social robots serve as caregivers and as toys, for example in shopping mall nurseries. Some also function as consulting and sales robots. Only in a few cases are they responsible for the tasks mentioned above, such as transportation and security. Selected examples are given below.

Several Pepper robots have been deployed in Zurich's Glatt shopping mall since 2017 (Vontobel and Weinmann 2017). They welcome customers, who can approach them and ask for information. If one of the models is overwhelmed by this interaction, an employee is switched on via its display on the chest to provide the requested information.

At MediaMarkt and Saturn, several models were trialled, such as Pepper and NAO and particularly, Paul (Dinske 2018). This is basically a Care-O-bot from the Fraunhofer Institute for Manufacturing Engineering and Automation IPA in Germany, originally intended for nursing and care, except the arms were removed. Paul was active in the Sihlcity shopping mall in Zurich, among other places.

Early on, California and Japan also experimented with social robots in retail. As of 2016, a Pepper was available to answer questions at the Westfield San Francisco Centre (Evangelista 2016). In Japan, it also popped up in several stores shortly after its "birth". The reports on this repeatedly sounded alarmist: "Tokyo firm replaces staff with a team of Pepper the 'emotional' humanoids." (Woollaston 2016) – so claimed *Mail Online* on March 24, 2016.

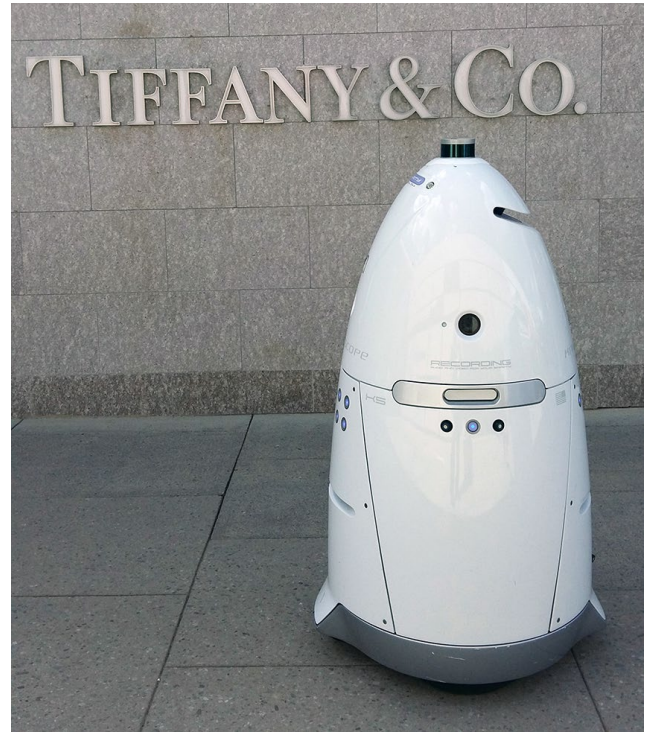

Fig. 1: K5 (Stanford Shopping Center, 2016)

A kangaroo-like robot has an unusual use in Tokyo. "In August, a robot vaguely resembling a kangaroo will begin stacking sandwiches, drinks and ready meals on shelves at a Japanese convenience store in a test its maker, Telexistence, hopes will help trigger a wave of retail automation." (Kelly 2020) Whether it is to be understood as a social robot in the narrower sense would have to be examined more closely. Without any doubt, it is a service robot.

## Functions of Social Robots in Retail

The social robot in retail uses its appearance to attract the customer's attention and to entertain him or her. It often has a humanoid design, at least in the form of a larger body and two arms. Often they will have eyes and a mouth that change color or move. The face is particularly important for customer engagement (Song and Luximon 2021). Forms of robot enhancement are possible, such as donning a uniform or wig.

Body and arm movements of the social robot appear entertaining and can encourage imitation – think of the dancing movements of NAO or the therapeutic movements of P-Rob or Lio by F&P Robotics in Switzerland (Bendel 2018). In addition, they serve as nonverbal communication. Customers appreciate being able to do the fist bump or high five with Pepper. Prasad et al. (2021) refer to this as human-robot handshaking. Its hand can hardly grasp anything – but is designed very naturally so that the greeting also appears natural (see Fig. 2).

Natural language capabilities are important to enable intuitive operation and to speak to the customer as a human consultant or salesperson would. The voice is crucial here, and it should sound pleasant and convincing. This is a challenge for robots like Pepper, which inherently have a robotic, childlike voice. Speech models such as GPT-2 and GPT-3, used in social robots such as Harmony, allow for longer conversations (Coursey 2020). Whichever speech model is used, they must be furnished with company-specific coordinates, and a knowledge base with appropriate data about the location and availability of products.

Some models used in retail are capable of face and voice recognition as well as gesture recognition, and some are capable of emotion recognition. Face and voice recognition can be used to identify a person so that the social robot can remember a customer or even, in conjunction with appropriate data, name them. It may also be used to determine age and gender, which can be important in the sales process. Emotion recognition provides information about the state of the customer upon entering the store, while shopping and receiving advice, and finally when leaving the store.

With the help of such means, it is also possible to categorize and select customers. It is possible to ask them about their wishes during a conversation and then make corresponding suggestions, or to classify and assign them on the basis of the automatically recognized age, gender, body shape, and state of mind, for example to relevant discount campaigns and special offers. If the robot were mobile or movable, it could take the customer to the checkout or at least show them the way – for security reasons, most social robots are found in a static (additionally secured) location in the retail store or shopping mall.

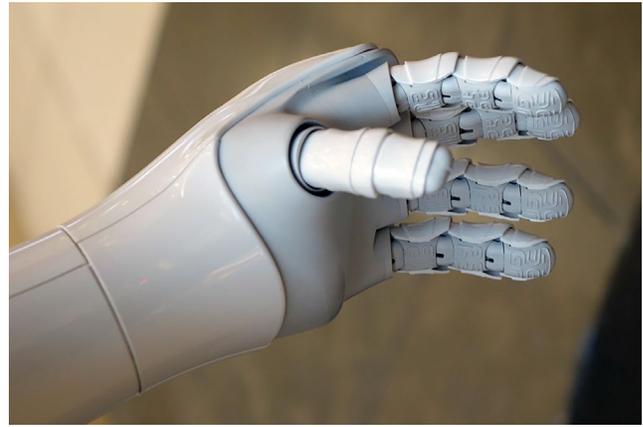

Fig. 2: Pepper's Hand (Westfield, 2017)

Another function is AI-supported analysis during the fitting of clothes. The system can, for instance, assess the fit of the garments and the matching of colors. It can also assist with a virtual fitting – so far, this has mainly been important in online retail, i.e. in situations where the garments are not physically available for a fitting (Werdayani and Widiaty 2021). In this context, the virtual fitting and analysis of the physical fitting could be combined.

In principle, video presentations, text, and image information are transmitted via an integrated display or via natural language. This can be classically acquired data, but also data acquired through machine learning and deep learning capabilities. Thus, in conjunction with appropriate AI systems, the social robot can learn from conversations with customers and from their behavior and apply this to new contacts. If it is connected to other systems, such as the booking system, it will also receive information about the success of its strategies and, in the best case, be able to adapt them itself.

## Should Social Robots Manipulate Customers?

Nowadays, social robots and especially humanoid variants are being used for several purposes in the education system, therapy and care, entertainment, hotel business, and retail sectors (Alves 2021). In doing so, these robots are becoming increasingly intelligent and their "behavior" less and less distinguishable from humans' behavior, making them well suited for consulting and sales assistance in retail stores.

Perhaps it is only a matter of time before the technology will reach a point when people will no longer deal with human advisors and sellers in retail stores but with humanoid robots – if stationary retail still has a chance at all. This is also supported by the advantages of automation listed at the beginning. Of course, there are also disadvantages, such as the lack of genuine social contact. However, according to several studies, robots are expected to be able to advise and sell at least as well or even better than human beings by 2025. It is expected that they will be more empathetic, more situation-specific, more flexible, more sophisticated, and more versatile. In short: more successful (Scheible 2019).

Knowing that human advisors and salespeople can manipulate customers, even though an international code of ethics for sales and marketing (Sirgy 2014) exists, albeit very succinctly and not everyone adheres to it, the question arises whether a social robot could also manipulate customers in retail stores to obtain an advantage. Manipulation here means that it directs their intentions so that, in extreme cases, they buy something they don't even want or need.

**Study about Manipulative Robots**

In her master thesis at the School of Business FHNW, Liliana Alves conducted a study on manipulative consulting and sales robots. The purpose of the study was both to provide transparency in the area of negative manipulation by humanoid robots and to fill the gap in ethical considerations of customer manipulation by humanoid consulting and sale assistance robots in retail stores (Alves 2021).

The main research question (RQ) was to determine whether it is ethical to intentionally program humanoid consulting and sales robots with manipulation techniques to influence the customer's purchase decision in retail stores. Moreover, to answer this central question, five sub-questions (SQ) were defined and answered based on an extensive literature review and a survey conducted with potential customers of all ages and varying socio-demographic characteristics. For SQ1, the goal was to find out how humanoid consulting and sales robots can manipulate customers in retail stores. Thereby, it was identified that social and humanoid robots can be programmed with manipulation content and are technically capable of manipulating customers, similar to human advisors or salespeople. As already indicated, there are several ways to do this, namely through vocal pitch, pacing of speech, voice volume, sentence melody, articulation, tone, words, semantics (e.g., questioning techniques, content-based manipulation concepts, argumentation concepts etc.), linguistic particularities, technical terms, foreign languages, posture, movement, gestures, facial expressions, and robot enhancement.

SQ2 aimed to determine if there are already ethical guidelines and policies to prevent humanoid robots from manipulating customers' purchasing decisions in the retail sector that developers and robot users must adhere to. Here, it was determined that manipulation is a known issue among various leading players (e.g., professional associations, national federations, and industry) in Europe, which is why some ethical guidelines and principles have already been created to avoid manipulation. However, manipulation is not explicitly related to social and humanoid robots, and these guidelines have not been developed specifically for retail sectors but for industries in general. Moreover, most of the guidelines and regulations are vaguely defined, so there are many ways to circumvent them and there are no subsequent sanctions if one does not adhere to the guidelines.

For SQ3, the goal was to find out if ethical guidelines and policies were established about who must perform the final inspection of the robots before they are placed into service. In this aspect, partially created ethical guidelines and policies were also disclosed. However, these tests or inspections do not explicitly refer to social and humanoid robots in retail stores. It is not specified which tests or inspections should be performed, and it is not mentioned who should or must complete the final tests or inspections before deploying robots in retail stores. Once again, the guidelines and regulations are very vaguely defined, and no sanctions exist.

With SQ4, the aim was to find out how potential customers in shopping malls and stores react, what they think and feel, when confronted with a manipulative humanoid advisor or sales robot in the retail sector. Thereby, it was identified that different thoughts, feelings, and reactions exist towards manipulative robots in retail stores. In fact, some people have an utterly negative opinion towards manipulative robots, others are neutral, and others are even positive. Yet, it can be concluded that generally, people do not want to be manipulated by humanoid robots and would therefore rather avoid these kinds of robots in the future or be more cautious when interacting with them.

Lastly, SQ5 should find out if potential customers accepted a manipulative and humanoid advisor or sales robot in a retail store. In this context, it was investigated whether customers might accept a manipulative robot in a retail store, but only if the manipulation is used positively by enhancing the customer's well-being or shopping experience. If, on the other hand, the manipulation is used to negatively influence the customer, it becomes neither acceptable nor ethically justifiable. The survey that produced these findings on SQ4 and SQ5 will now be looked at in more detail.

**Online Survey on Robot Manipulation**

The co-author conducted an online survey between February 9 and March 14, 2021. It was accessed 751 times, whereof 328 participants completed the survey (completion rate of 43.8 %). Approximately two-thirds were male, one-third female. All age groups were represented, with the largest numbers being 26- to 35-year-olds (109), 36- to 45-year-olds (79), and 46- to 55-year-olds (66), followed by 56- to 65-year-olds (41) (Alves 2021). In the evaluation of the survey no distinction was made between gender or age.

Two practical cases were presented in the online survey in order to answer SQ4 and SQ5. In the first, Pepper directly informs a customer that it has manipulated them. In the second, it does not inform the customer directly, and the customer discovers in an indirect way that they have been manipulated by the robot. In both cases, the assumed manipulation occurred through a specific form of content and tactical strategies that are also often used by human consultants and salespeople. Overall, it is primarily design, dialogue capability, and information transfer that are affected.

In the first case, around 108 people would be surprised that they have been manipulated, 93 people would feel deceived or betrayed, and 91 people would be upset. In the second case, 143 people said they would feel deceived or betrayed, 139 people would be upset, and 93 people would be surprised. In practice, the second case would be most likely to occur in real life scenarios. These two cases illustrate that customers would most probably feel deceived or betrayed and upset when being manipulated by a robot.

Summarizing the survey's main findings, the majority of the 328 participants prefer to be advised by a human advisor in a retail store (Alves 2021). Generally speaking, spontaneous advice is accepted, but this depends on the situation and the cordiality of the advisor or salesperson. Around 260 participants have no prejudices against human advisors or sellers. However, those who have prejudices believe that human advisors or sellers only want to sell without focusing on the customer's needs. Namely, they mainly aim to fulfil their own sales quota and sales targets, and to maximize profits to earn additional commissions and bonuses. A further prejudice is the lack of knowledge of the products or services they sell. Participants expect or believe that they often experience manipulation by humans during the retail store's counselling or sales process. Thereby, they feel either negatively or neutral when they become aware that they have been manipulated.

More than half of the participants are aware of the robot Pepper, and that manipulative robots exist in theory or practice. However, so far, only a small number of people (56) have had direct interaction with Pepper, and among these only ten people have ever received a consultation from it. Up to now, the experience with Pepper was generally rated as neutral by the participants.

When the survey confronted the participants with two hypothetical cases in which a manipulative robot negatively influenced them, about half of the participants stated that they had not expected such a situation and were initially somewhat surprised, amazed, and speechless. They would primarily doubt themselves as well as their purchase, and subsequently, they would probably feel angry, irritated, deceived, cheated, and unpleasant.

However, other participants viewed this neutrally and assumed that manipulation can happen anywhere. In this context, the participants believe that it depends on the customer whether they make the purchase, since there is no force applied. Ultimately, the robots are programmed by humans, which should make situations like this predictable. Furthermore, both robots and humans have the same task to fulfil, and it would be naive to believe that the robots are only being used in the customers' interests. Other participants indicated that they would even celebrate the robot, congratulate it and find the situation amusing. The participants mentioned that they would be grateful to learn to be more careful in the future and listen more to their intuition.

Most participants would probably want to return the purchase and receive a refund. Most would share their experiences with their personal and professional environment (e.g., via social media). Some participants mentioned that they would avoid such social robots in the future. Others would probably not enter the same retail store or never seek advice from a system like that again. However, other participants noted that they would simply be more careful when interacting with a robot in the future.

What was clear was that most participants do not want and do not accept manipulative robots in retail stores and further believe that society should not accept them either. For most people, it is not ethically justifiable to use manipulative robots in this context. According to these participants, robots in retail stores should be regulated for the purpose of "negative manipulation", but not banned from operating altogether. This is merely to say that the harmful facets of manipulation should be regulated. Instead, it is more important that a society becomes educated about such harmful actions to make responsible decisions.

Here, too, not all participants saw the situation the same way. For a smaller number it seemed appropriate for such manipulative robots to be used in retail stores. In fact, they stated that these robots are morally acceptable and should also be approved by society because they perform the same actions and tactics as humans. Thus, it is the customer's choice whether to be manipulated by them or not. As long as humans are allowed to influence customers negatively, robots should be allowed to do it, too, primarily because humans program them. For this reason, in these participants' opinion, these robots are ethically justifiable. Overall, the participants still prefer human interactions and would rather avoid the humanoid robot in retail altogether.

**Interim Conclusion**

The study, not relying only on the survey, can answer the main research question (RQ) as follows (Alves 2021): It is neither ethical for software developers to program robots with manipulative content nor is it ethical for companies to actively use these kinds of robots in retail stores to systematically manipulate customers in order to obtain an advantage. Business is about reciprocity, and it is not acceptable to systematically deceive, exploit, or manipulate customers to attain any kind of benefit.

However, it turns out that some survey participants find a manipulating robot acceptable or at least entertaining and amusing. Some also believe that humanoid or social robots should be allowed to cheat as long as humans do. This result must be taken seriously. Even if one were to find a different overall picture in further (larger and more representative) surveys, the individual statements would still remain. They will not be overly considered in the following section but they will be taken into account.

# Recommendations for Companies

The following section provides recommendations for companies looking to deploy social robots in retail settings. Some recommendations are of a general nature or are derived from previous experience in this area, while others are based on the findings of this paper and in particular on the evaluation of the study.

## Technical Perspective

It is fundamentally a decision for a company to employ a robot that supplements or replaces an employee. It is also fundamentally their decision to use a classic service robot without or with only a few social skills or a social robot with service skills. In doing so, the possibilities of AI can also be considered to a greater or lesser extent.

From a technical perspective, it is first important to ensure the functionality and security of the systems, i.e., the robot itself and the systems connected to it, such as databases and AI systems used for face recognition. This serves to establish fundamental trust in this type of technology and in this form of service.

Furthermore, a solid database must be guaranteed. The robot should know all the products in question and be able to name the prices and discounts correctly. Knowledge about the company itself and its customers is also important. When it comes to "world knowledge", many conversational agents and also social robots access Wikipedia, although it is not always reliable. Here, too, an alternative should be considered, even if it is merely Wikipedia articles that have been additionally reviewed (by the company's own experts).

It is also possible to offer customers various technical choices. For example, at the beginning of the consultation, they could select via a menu on the display whether the social robot should act more in the guise of a neutral sales system or that of a salesperson in the spirit of the MOME, the morality menu (Bendel 2020). Other aspects, such as the voice (female, male, or neutral) and the personality (serious, casual, funny, etc.) could also be selected.

Last but not least, it would be possible to let the user select or limit the AI-based systems individually. He or she could, for instance, do without facial recognition and related emotion recognition, thus protecting his or her privacy and informational autonomy. However, this would entail accepting restrictions in the shopping experience. They must also be prepared to be informed about the opportunities and risks, and be able to understand them.

## Ethical Perspective

In principle, it can be considered whether a social robot or a service robot in retail must have a humanoid design. When Pepper looks at the customer with its big eyes, it generates emotions in him or her, which it can recognize and reflect in its behavior and speech (see Fig. 3). In this context, one can certainly imagine a capacity for deception and fraud. At least users are made to believe that they have something alive in front of them, and they are manipulated in a certain way as one exploits their evolutionary tendency to react to something alive in a particular manner.

The operator should safeguard the use of social robots via ethical guidelines, similar to what the developer may have done previously during programming. These can be adopted from relevant initiatives and government agencies (in Europe, the High-level expert group on artificial intelligence should be mentioned, see Veale 2020) but should be adapted to the operator's own practice. It is important that the ethical guidelines are concrete, i.e., useful and implementable. In addition, non-compliance should result in sanctions. Legal provisions should also be taken into account, especially with regard to data protection and transparency.

In particular, the bias discussion should be considered during development and operation. The social or humanoid robot should not show any prejudices toward customers, just like consultants and salespeople. From this, there are behaviors that would be fundamentally prohibited, such as, for instance, negative statements made in the advisory and sales conversation with regard to age, gender, and ethnicity.

Prejudices and distortions should also be avoided through the use of data. This point is related to the technical perspective. The data and algorithms must be checked for biases, and actual biases must be removed. However, this will not always be possible, and correlations can prove useful for grouping and identifying potential customer preferences.

That many customers do not want to be manipulated must be taken into account. In this respect, it would tend to be easier to implement this in a robot than a consultant or salesperson, which is again related to the technical perspective. However, some respondents in the study stated that they had no problem with being manipulated, partly because this could contribute to their shopping experience. This could also be covered by a choice option, which again has to do with the technical perspective.

Some survey participants seem to think that social robots should be allowed to manipulate when humans are. This could be from a kind of sense of justice, although robots and humans are fundamentally different entities (which is not necessarily recognized by the users, especially since they are social robots that deliberately blur the differences). It was noted, however, that it is the programmer (together with other parties) who commands the manipulation, so to speak. The robot is simply reproducing human reality.

It could be the case that despite the neutral design or the possibility of choice on the part of the customers, manipulations of the robot arise, whether intended or unintended. If, despite assurances to the contrary, certain sales tricks and attempts at outwitting were used, this would be considered a breach of trust and critical from an ethical perspective.

However – as the survey also revealed – a robot that is not trustworthy could, in a certain sense, be an opportunity. Because if we become suspicious, we are less likely to fall for tricks of all kinds. This principle of experience also applies to technical systems. Responsible and trustworthy artificial intelligence is an important goal but educating users to its opposite seems to be just as important.

Last but not least, it must be noted that a store or shopping mall is not particularly the place for unembellished truth. It contributes to a customer's well-being to be flattered and complimented. Incidentally, he or she also does not want to be constantly told the truth about his or her appearance or behavior outside of shopping, especially if he or she does not conform to social expectations in their appearance.

### Economic Perspective

By using humanoid or social robots, a company can try to revitalise or increase the attractiveness of brick-and-mortar retail. It can demonstrate its willingness to transform itself, its business, and its innovative strength, thus setting itself apart from the competition. That being said, the novelty effect could quickly wear off.

A stationary retail company benefits from automation and AI methods. It can place its workforce in other areas or lay off workers, thereby reducing personnel costs (which raises ethical questions) and get to know its customers better. It can build a valuable data base based on the conversations and (re-)actions of its customers to better advise and assist groups or individuals, even those who do not fit the standard customer type.

Admittedly, some advisors and salespeople will see the social robot as direct competition, replacing them in their core activities. Meyer at al. surveyed "frontline employees" (FLEs) in a study: "The findings extend prior studies on technology acceptance and resistance and reveal […] that FLEs perceive service robots as both a threat and potential support. Moreover, they feel hardly involved in the co-creation of a service robot, although they are willing to contribute." (Meyer et al. 2020b)

If a company used manipulative social robots, this could damage its reputation. Even if it designed the system neutrally or granted a choice, it is not immune to criticism, especially if it turns out that manipulations were nevertheless present. Competitors can strike back here, so to speak, especially those retailers who reject the use of social robots for certain reasons, e.g., because they value social contacts.

It is also worth asking whether certain manipulations are not simply part of the business – what some of the interviewees believed. Advertising makes many promises that can hardly be kept, but which promote sales, and certain strategies of the sales staff also increase sales. It has become a game whose limits are constantly being tested. Just as some like to watch advertising that uses exaggeration, some might like to face social robots that try to trick them.

However, too much manipulation, whether it comes from the consultant or salesperson or from the social robot, could result in the original goal of reinforcing physical locality being reversed. Those who do not feel comfortable in the store, who are deceived and cheated, will ultimately prefer e-commerce – or send others to do the shopping.

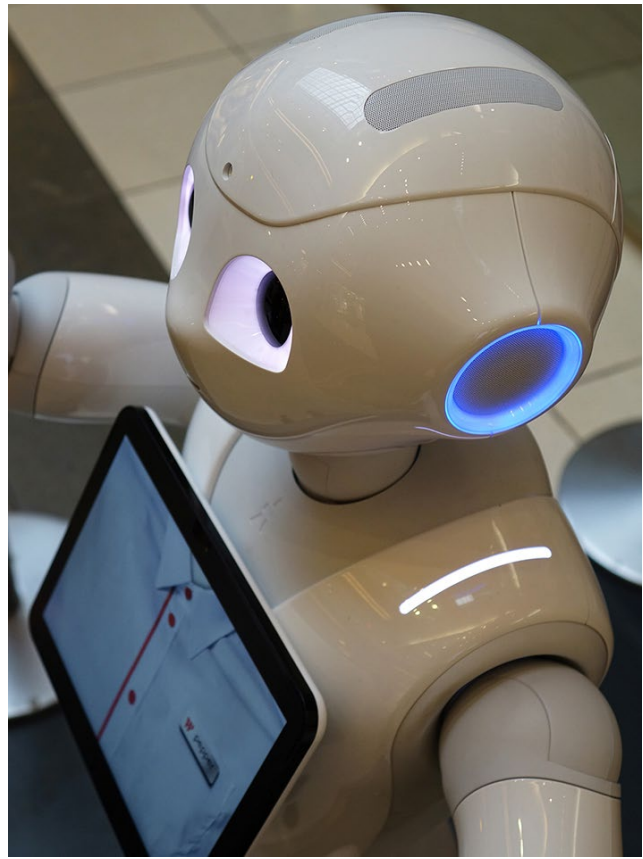

Fig. 3: Pepper (Westfield, 2017)

The economic perspective is actually a much broader one. When retailers employ social robots that obviously deceive and cheat, this shapes the image of other social robots that are urgently needed in service or care and therapy. This harms their manufacturers and the organizations and individuals who rely on their use. From this point of view, it would be economically and ultimately also ethically advisable to keep manipulation in this area low.

## Summary and Outlook

It can be assumed that social robots in the form of consulting and sales robots will play an increasingly important role in retail in the future as their capabilities grow because they are attractive to customers, the advisory service can be individualized and simultaneously standardized, and the sales process leading to a deal can be at least partially automated. These developments call for a scientific and sales-focused practical examination of buyer-robot interactions. In this context, the optimization of movement and natural language capabilities are central. The social robot is at the customer's side, in dialogue with them, listening to their questions and giving answers in different languages.

If social and specifically humanoid robots like Pepper (whose production was discontinued in 2020, which raises certain questions), NAO, Paul, and Cruzr have found their way into shopping stores and malls to attract, greet, and welcome customers, to inform, advise, and in the future persuade them to buy – should they behave like human advisors and salespeople, i.e. manipulate customers? Or should they be more honest and reliable than humans are? This article explored this question.

The conclusion is that manipulative behavior will please a small number of customers. They are interested in the game and the ambiguity or see a kind of equality in this possibility. The majority of customers, however, are likely to be interested in not being manipulated, and in a robot ultimately being less manipulative than a human advisor or salesperson. Ultimately, fair, honest, and trustworthy robots in retail are a win-win for everyone, not least for the company and yet, customers can be given choices that serve their usual or desired individual shopping experience. So how fair is fair? This depends on the wishes of the customers, but most of them expect to be served honestly and transparently by a social robot.

## References

Aaltonen, I.; Arvola, A.; and Heikkilä, P. 2017. Hello Pepper, May I Tickle You?: Children's and Adults' Responses to an Entertainment Robot at a Shopping Mall. In *HRI '17: Proceedings of the Companion of the 2017 ACM/IEEE International Conference on Human-Robot Interaction*, March 2017, 53–54.

Alves, L. 2021. *Manipulation by humanoid consulting and sales hardware robots from an ethical perspective*. Master Thesis. Olten: School of Business FHNW.

Bendel, O. (ed.). 2021a. *Soziale Roboter: Technikwissenschaftliche, wirtschaftswissenschaftliche, philosophische, psychologische und soziologische Grundlagen*. Wiesbaden: Springer Gabler.

Bendel, O. 2021b. Das steht Ihnen aber gut!: Empfangs-, Beratungs- und Verkaufsroboter im Detailhandel. In Bendel, O. (ed.) *Soziale Roboter: Technikwissenschaftliche, wirtschaftswissenschaftliche, philosophische, psychologische und soziologische Grundlagen*. Wiesbaden: Springer Gabler, 495–515.

Bendel, O. 2020. The Morality Menu Project. In Nørskov, M.; Seibt, J.; Quick, O. S. (eds.) *Culturally Sustainable Social Robotics – Challenges, Methods and Solutions: Proceedings of Robophilosophy 2020*. IOS Press, Amsterdam, 257–268.

Bendel, O. (ed.) 2018. *Pflegeroboter*. Wiesbaden: Springer Gabler.

Bendel, O.; Schwegler, K.; and Richards, B. 2017. Towards Kant Machines. In *The 2017 AAAI Spring Symposium Series*. AAAI Press, Palo Alto 2017, 7–11.

Coursey, K. 2020. Speaking with Harmony: Finding the right thing to do or say ... while in bed (or anywhere else). In Bendel, O. (ed.) *Maschinenliebe: Liebespuppen und Sexroboter aus technischer, psychologischer und philosophischer Sicht*. Wiesbaden: Springer Gabler, 35–51.

Evangelista, B. 2016. Robots greet Westfield mall shoppers in San Francisco, San Jose. *SFGATE*, 22 November 2016. https://www.sfgate.com/business/article/Robots-greet-Westfield-mall-shoppers-in-San-10631291.php.

Kelly, T. 2020. Japanese robot to clock in at a convenience store in test of retail automation. 15 July 2020. In *Reuters.com*, https://www.reuters.com/article/us-japan-tech-robot-idUSKCN24G138.

Meyer, P.; Jonas, J. M.; and Roth, A. 2020a. Exploring customer's acceptance of and resistance to service robots in stationary retail – a mixed method approach. In *ECIS 2020 Proceedings at AIS Electronic Library* (AISeL). Research Papers, 9, https://aisel.aisnet.org/ecis2020_rp/9.

Meyer, P.; Jonas, J. M.; and Roth, A. 2020b. Frontline Employees' Acceptance of and Resistance to Service Robots in Stationary Retail: An Exploratory Interview Study. *SMR Journal of Service Management Research*, Volume 4, 1/2020:21–34.

Prasad, V.; Stock-Homburg, R.; and Peters, J. 2021. Human-Robot Handshaking: A Review. In *arXiv.org*, 14 February 2021, https://arxiv.org/abs/2102.07193.

Scheible, K.-G. 2019. Roboter schlägt Mensch – Verhandlungen der Zukunft. In Buchenau, P. (ed.) *Chefsache Zukunft: Was Führungskräfte von morgen brauchen*. Wiesbaden: Springer Gabler, 507–520.

Sirgy, M. J. 2014. *Real Estate Marketing: Strategy, Personal Selling, Negotiation, Management, and Ethics*. London: Taylor & Francis.

Song, Y.; and Luximon, Y. 2021. The face of trust: The effect of robot face ratio on consumer preference. *Computers in Human Behavior*, Volume 116, March 2021, 106620. https://www.sciencedirect.com/science/article/pii/S0747563220303678.

Veale, M. 2020. A Critical Take on the Policy Recommendations of the EU High-Level Expert Group on Artificial Intelligence. *European Journal of Risk Regulation*:1–10.

Vontobel, N.; and Weinmann, B. 2017. So motzt die Migros Konsumtempel im Kampf gegen Onlineshops auf. *Tagblatt*, 7 October 2017. https://www.tagblatt.ch/wirtschaft/so-motzt-die-migros-konsumtempel-im-kampf-gegen-onlineshops-auf-ld.1456998.

Werdayani, D.; and Widiaty, I. 2021. Virtual fitting room technology in fashion design. In *IOP Conf. Ser.: Mater.* Sci. Eng. 1098 022110.

Woollaston, V. 2016. The phone store run by ROBOTS: Tokyo firm replaces staff with a team of Pepper the 'emotional' humanoids. *Mail Online*, 24 March 2016. https://www.dailymail.co.uk/sciencetech/article-3507541/The-phone-store-run-ROBOTS-Tokyo-firm-replaces-staff-10-versions-Pepper-emotional-humanoid.html.